\documentclass[final,5p,times,twocolumn]{elsarticle}

\usepackage{times}
\usepackage{epsfig}
\usepackage{graphicx}
\usepackage{subfigure}
\usepackage{amsmath}
\usepackage{amssymb}
\usepackage[numbers]{natbib}
\usepackage{booktabs}  
\usepackage{multirow}
\usepackage{setspace}
\usepackage{floatrow}

\begin{document}

\begin{frontmatter}
\biboptions{sort&compress}
\title{A Dense Siamese U-Net trained with Edge Enhanced 3D IOU Loss for Image Co-segmentation}


\author[]{Xi Liu}
\author[]{Xiabi Liu}
 
\author[]{Huiyu Li}
\author[]{Xiaopeng Gong}
\address[]{Beijing Lab of Intelligent Information Technology, School of Computer Science, Beijing Institute of Technology, Beijing 100081, China}

\cortext[cor1]{Corresponding author}

\begin{abstract}
Image co-segmentation has attracted a lot of attentions in computer vision community. In this paper, we propose a new approach to image co-segmentation through introducing the dense connections into the decoder path of Siamese U-net and presenting a new edge enhanced 3D IOU loss measured over distance maps. Considering the rigorous mapping between the signed normalized distance map (SNDM) and the binary segmentation mask, we estimate the SNDMs directly from original images and use them to determine the segmentation results. We apply the Siamese U-net for solving this problem and improve its effectiveness by densely connecting each layer with subsequent layers in the decoder path. Furthermore, a new learning loss is designed to measure the 3D intersection over union (IOU) between the generated SNDMs and the labeled SNDMs. The experimental results on commonly used datasets for image co-segmentation demonstrate the effectiveness of our presented dense structure and edge enhanced 3D IOU loss of SNDM. To our best knowledge, they lead to the state-of-the-art performance on the Internet and iCoseg datasets.
\end{abstract}



\end{frontmatter}


\section{Introduction}
Image co-segmentation is a problem of segmenting common and salient objects from a set of related images. Since this concept was firstly introduced in 2006 ~\cite{b1}, it has attracted a lot of attentions. It has been widely used to support various computer vision applications, such as interactive image segmentation~\cite{b2}, 3D reconstruction ~\cite{b3}, object co-localization ~\cite{b4,b5}, and etc. 

Conventional co-segmentation approaches utilize handcrafted features and prior knowledge~\cite{b7,b8,b6}. They are difficult to achieve good robustness. In recent years, deep learning is introduced to learn visual representation in a data driven manner for improving the performance of image co-segmentation~\cite{b10,b11,b9}. It has shown promising results. However, the exact image co-segmentation is still far away from our expectation. We still face the challenges like background clutter, appearance variance of co-object across images, similarity between co-object and non-common object, and etc. These challenges especially result in unsatisfactory prediction along object edges.
 
In this work, we propose a new deep learning approach based on signed normalized distance map (SNDM) for improving the performance of image co-segmentation. The distance map is a special representation of segmentation masks, in which the values reflect the spatial proximity to the object boundary of each pixel and the sign of values indicates the segmentation result. We transform the segmentation problem to a SNDM regression problem and solve this regression problem through constructing a dense U-shaped Siamese network and learning it with an edge enhanced 3D IOU loss. The proposed approach is evaluated on commonly-used datasets for image co-segmentation.
Our main contributions are summarized as follows.

(1) We introduce the SNDM into image co-segmentation. The segmentation problem is transformed into a SNDM regression problem. Since SNDM contain much more plentiful information than binary segmentation mask, it is potential to help improve the segmentation results, especially in object boundaries.

(2)A new dense Siamese U-net neural network is presented to complete SNDM regression, in which the blocks in the decoder part are densely connected to utilize multi-scale features more sufficiently.

(3)A new edge enhanced 3D IOU loss over SNDM is proposed by taking a SNDM as a 3D shape and penalizing segmentation errors at object boundaries.

The rest of this paper is organized as follows. Section 2 reviews the related work. Section 3 presents our dense Siamese U-net and edge enhanced 3D IOU learning loss for training. The experimental results are reported in Section 4. We conclude in Section 5.

\section{Related work}

Our main contributions are mainly related with the distance map based image segmentation and dense connection. The corresponding previous works are briefly introduced as follows.

\subsection{Image segmentation based on distance map}

Most of image segmentation methods use binary or multi-label mask as ground truth. Distance map offers an alternative to classical ground truth. Incorporating the distance maps of image segmentation labels into convolutional neural network (CNN) pipelines has received significant attention. 

In Hayder et al.~\cite{b12}, a distance transform-based mask representation was introduced to allow an instance segmentation to predict shapes beyond the limits of initial bounding boxes, which allows the network to learn more specific information about the location of the object boundary than binary mask representation would do. Tan et al.~\cite{b13} added a decoder branch to do mask estimation and boundary distance map regression, where the distance map estimation acts as the supervisor to mask prediction. Dangi et al.~\cite{b14} proposed a multi-task learning based regularization framework to perform the main task of semantic segmentation, and an auxiliary task of pixel-wise distance map regression, simultaneously. Yin et al.~\cite{b15} modeled the kidney boundaries as boundary distance maps and predicted them in regression setting. Subsequently, the predicted boundary distance maps were used to learn pixelwise kidney masks.

Some methods use distance map to design new loss functions. Jia et al.~\cite{b16} employed a contour loss based on distance map information to obtain the segmentation of boundary regions. Caliva et al.~\cite{b17} used distance maps as the penalty term of the cross-entropy (CE) loss, enforcing the network to focus on the hard-to-segmented object boundaries. Boundary loss~\cite{b18} assigned the weights to the softmax probability outputs based on the ground truth distance map, while Hausdorff distance loss~\cite{b19} introduced not only the ground truth distance map but also the predicted segmentation distance map to weight the softmax probability outputs. SDF loss~\cite{b20} employed the product of predicted distance map and ground truth distance map to guide the distance map regression network during training. 
%

\subsection{Dense connection}

For deep CNNs, as information about the input or gradient passes through many layers, it can vanish and “wash out” by the time it reaches the end (or beginning) of the network. Many studies have demonstrated this or related problems and emphasize the importance of using feature of shallow layers to optimize features from deep layers~\cite{b21}. Huang et al.~\cite{b22} proposed DenseNet, which utilizes a dense connection method to cope with the vanishing gradients problem. In DenseNet, to preserve the feed-forward nature, each layer obtains additional inputs from all preceding layers and passes on its own feature-maps to all subsequent layers. To address the issue of preserving spatial information in the U-Net architecture, Dong et al.~\cite{b23} designed a dense feature fusion module using the back-projection feedback scheme. It shows that the dense feature fusion module can simultaneously remedy the missing spatial information from high-resolution features and exploit the non-adjacent features. Zhang et al.~\cite{b24} proposed an edge-preserving densely connected encoder-decoder structure with multilevel pyramid pooling module for estimating the transmission map for their single image dehazing method. 

Dense connection allows feature reuse throughout the networks and can consequently learn more compact and more accurate models. It alleviates the vanishing-gradient problem, strengthens feature propagation, and encourages feature reuse.

\section{The Proposed Method}

In this section, we present the details of our proposed method. First, SNDM is introduced in section 3.1. Then, our dense Siamese U-Net for SNDM regression is described in section 3.2. Finally, our edge enhanced 3D IOU loss of SNDM is presented in section 3.3.

\begin{figure}[htbp]
\centering
\subfigure[ ]{
\begin{minipage}[t]{0.33\linewidth}
\centering
\includegraphics[width=1in]{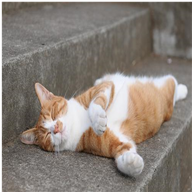}
\end{minipage}%
}%
\subfigure[ ]{
\begin{minipage}[t]{0.33\linewidth}
\centering
\includegraphics[width=1in]{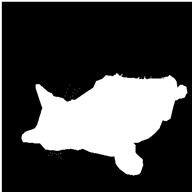}
\end{minipage}%
}%
\subfigure[ ]{
\begin{minipage}[t]{0.33\linewidth}
\centering
\includegraphics[width=1in]{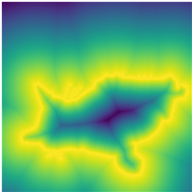}
\end{minipage}
}%
\centering
\caption{An example of SNDM: (a) the original image, (b) binary segmentation mask, and (c) SNDM generated on Fig. 1}
\label{fig1}
\end{figure}

\begin{figure*}[t]
\begin{center}
\includegraphics[width=0.7\linewidth]{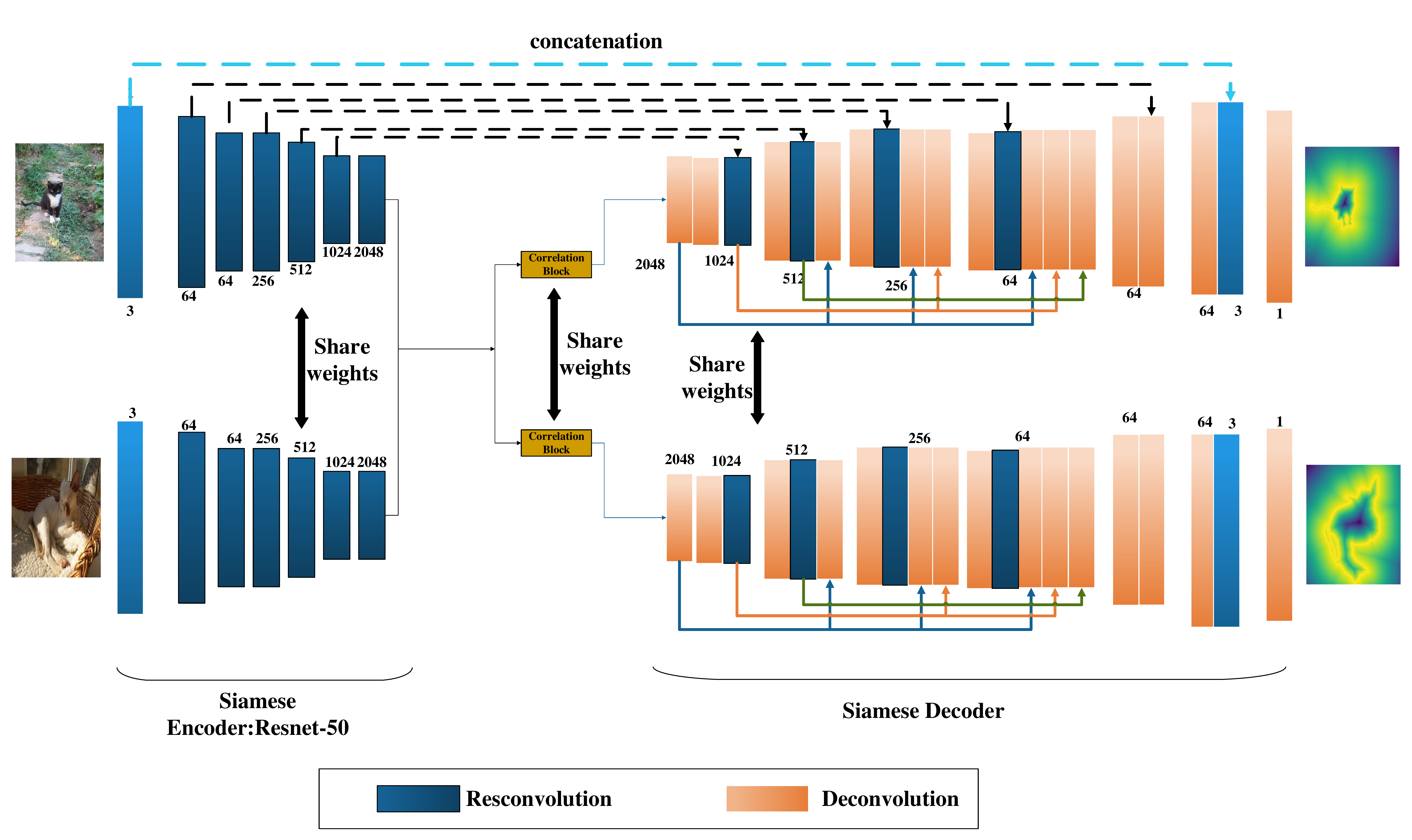}
\caption{The architecture of our dense Siamese U-net for image co-segmentation.}
\label{fig2}
\end{center}
\end{figure*}

\subsection{Signed Normalized Distance Map}
We generate a SNDM on the binary segmentation mask to obtain richer information about structural features of objects. For each pixel in a segmentation mask, we compute the Euclidean distance between this pixel and the closest pixel in the boundary of the target object as the original distance map. Then the values in the original distance map is normalized to be in the range [0, 1] by the local maximum distance~\cite{b25}. We further add the sign indicating foreground (positive) and background (negative). Let d(x, b) be the Euclidean distance between the point x and b, B be the set of points on the object boundary, then the SNDM is computed by using
\begin{footnotesize}
\begin{equation}
M(x) = \left\{ {\begin{array}{*{20}{c}}
{ - \frac{{\max (D(x)) + 1 - D(x)}}{{\max (D(x))}},x \in background}\\
{\frac{{\max (D(x)) + 1 - D(x)}}{{\max (D(x))}},x \in foreground}
\end{array}} \right.
\label{eq1}
\end{equation}
\end{footnotesize}
where
\begin{footnotesize}
\begin{equation}
D(x) = \mathop {\min (d(x,b))}\limits_{\forall b \in B}
\label{eq2}
\end{equation}
\end{footnotesize}

According to Eq.~\ref{eq1}, the value of SNDM for a pixel far from the object boundary is very close to 0, the weight for this pixel is low and the sign of it is easy to be predicted wrongly. For solving this problem, we perform a linear transformation on M(x) and normalized the transformed M(x) to [0.1,1] for foreground and [-1,0.1] for background. Fig.~\ref{fig1} shows an example of SNDM. 

\begin{figure}[t]
\centering
\includegraphics[width=0.8\linewidth]{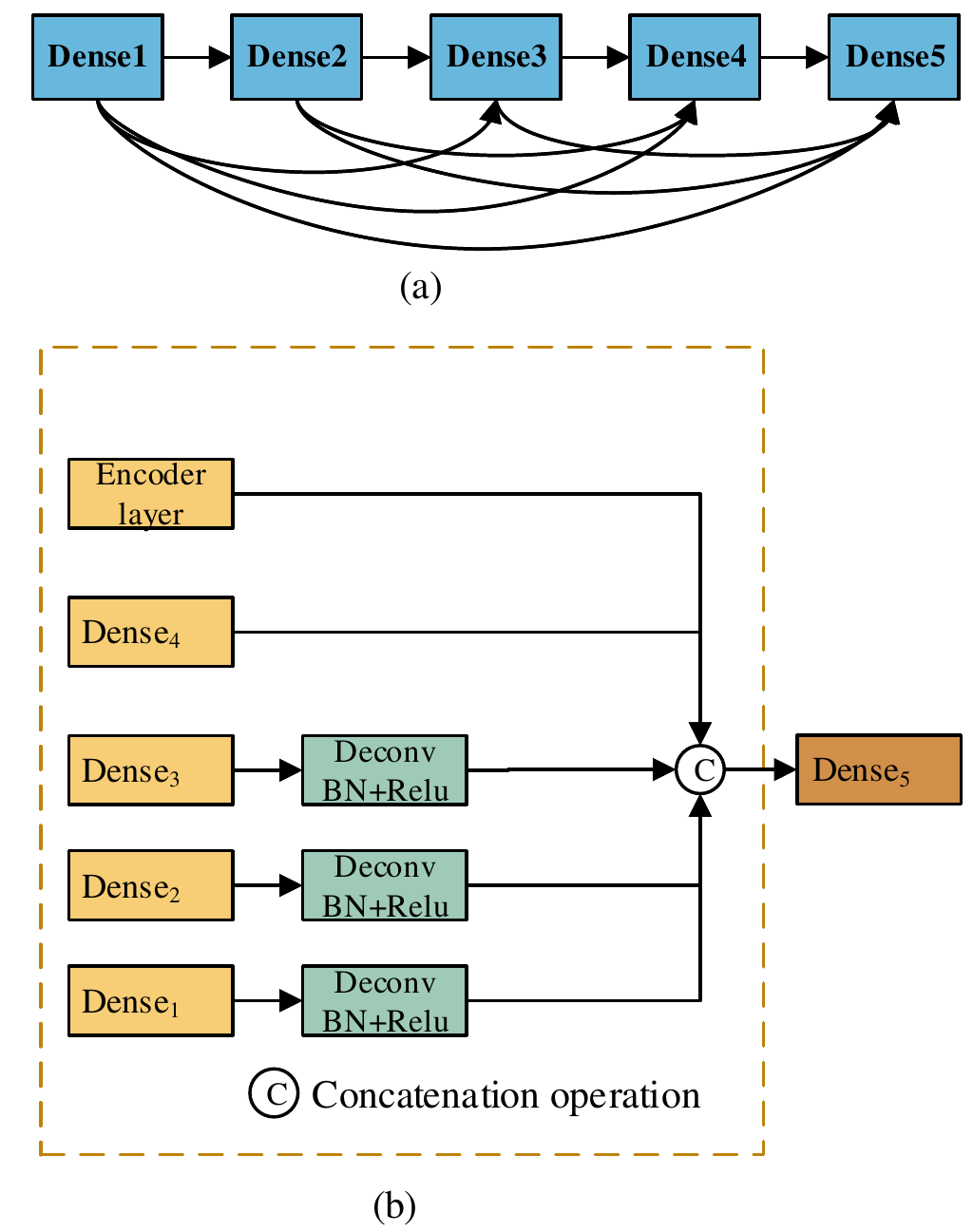}
\centering
\caption{The illustration of dense modules: (a) dense connections; (b) the operations in a dense connection.}
\label{fig3}
\end{figure}

In the SNDM, there is a mutation from -1 to 1 around the object boundary, which is helpful to strengthen the distinguish between the foreground and the background in blurry edges.

\begin{figure}[t]
\begin{center}
\includegraphics[width=1\linewidth]{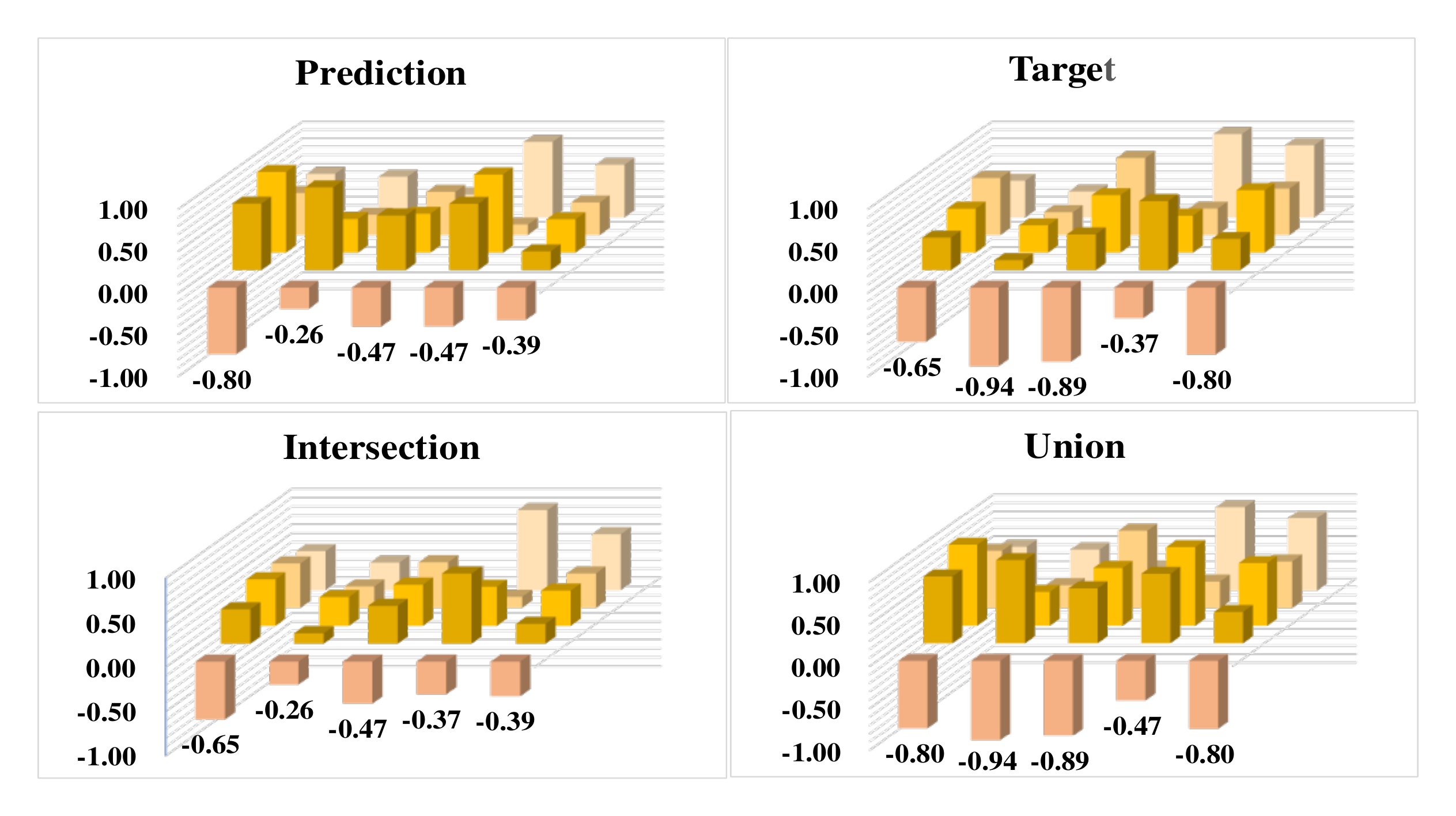}
\caption{The illustration of how to compute intersection and union between the predicted SNDM and the target SNDM.}
\label{fig4}
\end{center}
\end{figure}

\subsection{Network Architecture}

As shown in Fig.~\ref{fig2}, the overall structure of our segmentation network is a Siamese U-net enhanced by the dense connections in the decoder path, which is composed of three parts. The first part is a Siamese encoder which consists of two identical ResNet50 networks with shared parameters for feature extraction. The second part is the correlation block, through which the correlation maps are calculated from the two feature maps. The third part is a Siamese decoder network that consists of a hierarchy of decoders, each corresponding to an encoder layer. The values in SNDM are in the range of [-1,1], so we use Tanh activation function in the last decoder layer. Except this last decoder layer, ReLU activation function is used in all the other layers. Furthermore, we concatenate the input image in the last decoder layer for determining the result. This is ignored in previous U-net for image segmentation. 

As show in  Fig.~\ref{fig2}, the decoder subnetwork consists of multiple dense modules. The output of each module will be passed to all the following modules. This dense connection is illustrated in  Fig.~\ref{fig3} (a) and the computation of each dense module is illustrated in  Fig.~\ref{fig3} (b) by taking the final module as an example. As shown in  Fig.~\ref{fig3} (a), a dense module accepts the features from the corresponding encoder layer and the features from all the previous decoder layers. For those feature maps that have different spatial resolution with current module, we transform them to feature maps with the same spatial resolution by performing the computation composed of deconvolution, batch normalization and rectified linear unit (ReLU). Finally, all of the features are fused by the concatenation operation. 

\subsection{Edge Enhanced 3D IOU Loss of SNDM}
For training our network of regressing SNDM, we present a new loss by adapting the popular Dice loss~\cite{b26,b27} to our problem. Dice loss can be used for data sets with imbalanced positive and negative samples. It usually behaves better than the CE loss in segmentation applications. We can view a binary segmentation mask as a plane shape. Then the Dice loss is actually defined by the intersection and the union of two plane shapes corresponding to labeled segmentation and predicted segmentation, respectively. The values in a SNDM are in [-1, 1]. So, a SNDM cannot be seen as a plane shape, but a 3D shape. We illustrate such view of SNDM in Fig.~\ref{fig4}. Accordingly, we can calculate the intersection and the union of two 3D shapes to define the intersection over union(IOU) loss for two SNDMs. Let ${{g_i}}$ and ${{p_i}}$ be the value for the i-th foreground pixel in the labeled and the predicted SNDM, respectively, ${{u_j}}$ and ${{v_j}}$ be the value for the j-th background pixel in the labeled and the predicted SNDM, respectively, N and M be the number of foreground and background pixels, respectively, then the 3D IOU loss of SNDM is

\begin{footnotesize}
\begin{equation}
L = 1 - \frac{{\sum\nolimits_i^N {\min ({p_i},{g_i}) + \sum\nolimits_j^M {\min ( - {u_j}, - {v_j})} } }}{{\sum\nolimits_i^N {\max ({p_i},{g_i}) + \sum\nolimits_j^M {\max ( - {u_j}, - {v_j})} } }}
\label{eq3}
\end{equation}
\end{footnotesize}

For computing 3D IOU, we should use the absolute value measured in each pixel, so the values of  background pixels are negated in Eq.~\ref{eq3}.

Our final purpose is to obtain accurate segmentation results. The sign of values in SNDM indicate the segmentation results. Thus if the sign of a value in labeled SNDM is opposite with that of the corresponding value in predicted SNDM, a misclassification must occurr. According to this knowledge, we can extend our 3D IOU loss by imposing a penalty on this kind of errors to improve the segmentation accuracy. We compute this penalty by using

\begin{footnotesize}
\begin{equation}
F = \left\{ {\begin{array}{*{20}{c}}
{1,\quad {p_i}*{g_i} > 0}\\
{\lambda,\quad {p_i}*{g_i} <  = 0}
\end{array}} \right.
\label{eq4}
\end{equation}
\end{footnotesize}

and the loss is extended to

\begin{footnotesize}
\begin{equation}
L  = 1 - \frac{{\sum\nolimits_i^N {\min ({p_i},{g_i}){F_i} + \sum\nolimits_j^M {\min ( - {u_j}, - {v_j})} } {F_j}}}{{\sum\nolimits_i^N {\max ({p_i},{g_i}){F_i} + \sum\nolimits_j^M {\max ( - {u_j}, - {v_j}){F_j}} } }}
\label{eq5}
\end{equation}
\end{footnotesize}

$\lambda$ in Eq.~\ref{eq4} is a trade-off parameter for balancing the objective of segmentation accuracy and the objective of SNDM regression. To increases $\lambda$ increases the importance of correcting misclassified examples. The appropriate $\lambda$ needs to be determined in the applications. In our experiments, the best $\lambda$ is observed to be 5. 

Finally, the pixels closer to the object edge are more likely to be misclassified. Thus it is helpful to let the pixels closer to the object edge paid more attention in the training. Based on this idea, we add a weight to each pixel, which is proportional to the value of labeled SNDM. The final form of our loss is
\begin{footnotesize}
\begin{equation}
L = 1 - \frac{{\sum\nolimits_{\rm{i}}^N {\min ({p_i},} {g_i}){F_i}*\sqrt {{g_i}}  + \sum\nolimits_{\rm{j}}^M {\min ( - {u_j},}  - {v_j}){F_j}*\sqrt { - {u_j}} }}{{\sum\nolimits_{\rm{i}}^N {\max ({p_i},} {g_i}){F_i}*\sqrt {{g_i}}  + \sum\nolimits_{\rm{j}}^M {\max ( - {u_j},}  - {v_j}){F_j}*\sqrt { - {u_j}} }}
\label{eq6}
\end{equation}
\end{footnotesize}

Based on this loss, our dense Siamese U-net is optimized with the learning algorithm of Adam~\cite{b34}.

\section{Experiment}

\floatsetup[table]{capposition=top}
\newfloatcommand{capbtabbox}{table}[][0.9\FBwidth]
\begin{table*}[t]
	\caption{The performance comparisions in Internet, where bold indicate the best results.}\smallskip
	\centering
	\resizebox{.7\columnwidth}{!}{
		\smallskip\begin{tabular}{lllllllll}
			\toprule
			\multirow{2}{*}{Method} & \multicolumn{2}{l}{Airplane} & \multicolumn{2}{l}{Car} & \multicolumn{2}{l}{Horse} & \multicolumn{2}{l}{Average} \\
			\cmidrule(r){2-3} \cmidrule(r){4-5} \cmidrule(r){6-7}\cmidrule(r){8-9}
			&  Precision      &  Jaccard  &  Precision      &  Jaccard   &  Precision      &  Jaccard    &  Precision      &  Jaccard \\
			\midrule
			Jerripothula et al. \cite{b8}       &90.5 &0.61 &88.0 &0.71 &88.3 &0.60 &88.9 &0.64\\
			Han et al. \cite{b5}      &92.3 &0.60 &88.7 &0.68 &89.3 &0.58 &90.1 &0.62 \\
			Yuan et al. \cite{b37}     &92.6 &0.66 &90.4 &0.72 &90.2 &0.65 &91.1 &0.68\\
			Li et al. \cite{b9}      &94.6 &0.64 &94.0 &0.83 &91.4 &0.65 &93.3 &0.71\\
 			Chen et al.~\cite{chen2019show} &94.1 &0.65 &94.0 &0.82 &92.2 &0.63 &95.2 &0.78 \\
			Gong et al. ~\cite{b38} &95.5 &0.76 &94.7 &0.87 &93.3 &0.65 &94.5 &0.76\\
			OURS    & \textbf{97.0} &\textbf{0.81} &\textbf{96.3} &\textbf{0.90} &\textbf{93.7} &\textbf{0.74} &\textbf{95.7} &\textbf{0.82}\\
			\bottomrule
		\end{tabular}
	}
	\label{tab1}
\end{table*}

\begin{figure*}[t]
\begin{center}
\includegraphics[width=0.8\linewidth]{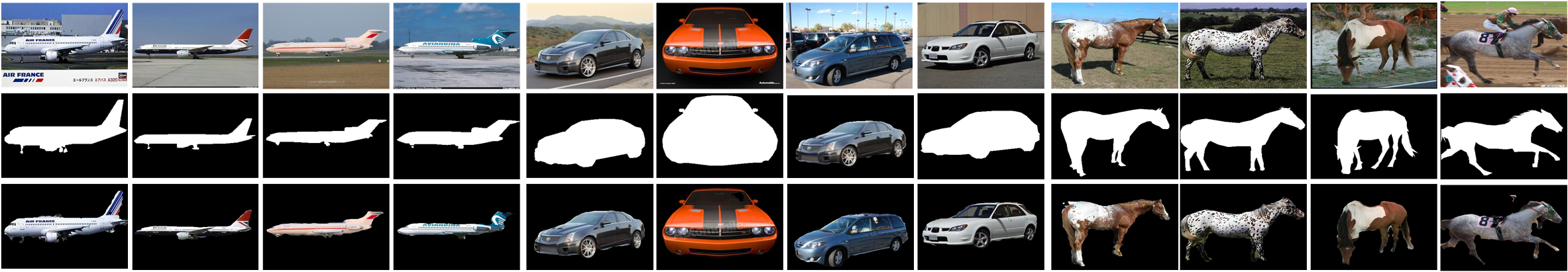}
\caption{The co-segment results generated by our approach on the Internet dataset: the examples of airplane, car and horse are shown in sequence from left to right; the input images, the ground-truths, and the segmented results are shown in the top, middle and bottom row, respectively.
iCoseg tests.}
\label{fig5}
\end{center}
\end{figure*}

\subsection{Experimental Setup}
\textbf{Datasets.}
Natural images and commodity images are tested. For natural images, we use Pascal VOC 2012 [28] and MSRC [29] datasets to train our image co-segmentation network, and then use Internet [30] and sub-set of iCoseg [7] as the test sets. These four data sets are widely used in the community of image co-segmentation. These datasets are composed of real-world images with large intraclass variations, occlusions and background clutters. MSRC is composed of 591 images of 21 object groups. The groundtruth is roughly labeled, which does not align exactly with the object boundaries. VOC 2012 includes 11,540 images with ground-truth detection boxes and 2913 images with segmentation masks. Only 2913 images with segmentation masks can be considered in our problem. Note that not all the examples in these two datasets can be used. In MSRC, some images include only stuff without obvious foreground, such as only sky or grassland. In VOC 2012, the interested objects in some images have great changes in appearance and are cluttered in many other objects, so the meaningful correlation between them is ambiguous. We exclude them from consideration. The remained 1743 images in VOC 2012 and 507 images in MSRC are used to construct our training and validation set. From the training images, we sampled 13200 pairs of images containing common objects as training set and 857 pairs of images as validation set to train our image co-segmentation network.

The Internet dataset contains images of three object categories including airplane, car and horse. Thousands of images in this dataset were collected from the Internet. Following the same setting of the previous work ~\cite{b10,b30,b31}, we use the same subset of the Internet dataset where 100 images per class are available. iCoseg consists of 38 groups of total 643 images, each group contains 17 images on average. It contains images with large variations of viewpoints and multiple co-occurring object instances and difficult segmentaion. Following the compared methods, we evaluate our approach on its widely used subset. 

For commodity image, we use SBCoseg dataset ~\cite{b32} that is a new challenging image dataset with simple background for evaluating co-segmentation algorithms. It contains 889 image groups with 18 images in each and the pixel-wise hand-annotated ground truths. The dataset is characterized by simple background produced from nearly a single color. It looks simple but is actually very challenging for current co-segmentation algorithms, as there are four difficult cases in it: easy-confused foreground with background(ECFB), transparent regions in objects(TP), minor holes in objects(MH), and shadows(SD). we divide the SBCoseg into 13842, 720, and 1440 images for training, validation, and testing, respectively. Each subset contains all the ECFB, TR, MH, SD, and normal cases.

Evaluation metrics. We use two commonly used metrics for evaluating the effects of image co-segmentation: Precision and Jaccard index ~\cite{b33}. 
Precision is the percentage of correctly classified pixels in both background and foreground, which can be defined as
\begin{footnotesize}
\begin{equation}
Precision= \frac{|Segmentation \cap Groundtruth|}{|Segmentation|}
\label{eq8}
\end{equation}
\end{footnotesize}

The background pixels are taken into account in precision, so the image with large background area and small foreground area tend to perform well in precision. Therefore, the precision may not be very faithful to evaluate the performance of algorithms. Jaccard index is used to compensate for this shortcoming. Jaccard index (denoted by Jaccard in the following descriptions) is the overlapping rate of foreground between the segmentation result and the ground truth mask, which can be defined as
\begin{footnotesize}
\begin{equation}
Jaccard{\rm{ = }}\frac{{Segmentation \cap Segmentation}}{{Segmentation \cup Groundtruth}}
\label{eq9}
\end{equation}
\end{footnotesize}

\textbf{Parameter setting.}
We conduct the experiments on a computer with GTX 1080Ti GPU and implement the image co-segmentation network with PyTorch. In the experiments, the batch size for training is set to be 4, the learning rate is initialized to 0.00001 and is divided by 2 as the loss in the validation data do not decrease for 10 epochs. The optimization procedure ends after 120 epochs. We use Adam optimize ~\cite{b34} and the weight decay is set to be 5e-5. Considering the limited computing resource, we resize the input image to the resolution of 512*512 in advance. The co-segmentation results are resized back to the original image resolution for performance evaluation.

 \begin{table*}[t]
	\caption{The comparisons of Jaccard index on iCoseg-subset with 8 groups of images. The bold indicate the best results among all methods.}\smallskip
	\centering
	\resizebox{.70\columnwidth}{!}{
		\smallskip\begin{tabular}{llllll}
			\toprule
			 Class & Faktor and Irani ~\cite{b35}& Jerripothula et al.~\cite{b8}& Li et al.~\cite{b9}& Gong et al.~\cite{b38}& Ours\\
			\midrule
			Bear2& 0.70& 0.68& 0.88& \textbf{0.89}& 0.87\\
			Brownbear& 0.92& 0.73& 0.92& \textbf{0.94}& \textbf{0.94}\\
			Cheetah& 0.67& 0.78& 0.69& 0.78& \textbf{0.89}\\
			Elephant& 0.67& 0.80& 0.85& 0.88& \textbf{0.89}\\
			Helicopter& 0.82& 0.80& 0.92& \textbf{0.94}& 0.88\\
			Hotballoon& 0.88& 0.80& 0.92& 0.94& \textbf{0.96}\\
			Panda1& 0.70& 0.72& 0.83& 0.86& \textbf{0.90}\\
			Panda2& 0.50& 0.61& 0.87& \textbf{0.88}& 0.87\\
			Average& 0.78& 0.74& 0.84& 0.87& \textbf{0.90}\\
			\bottomrule
		\end{tabular}
	}
	\label{tab2}
\end{table*}

\begin{table*}[t]
	\caption{The results on SBCoseg dataset. }
	\centering
	\resizebox{.95\columnwidth}{!}{
		\smallskip\begin{tabular}{lllllllllllll}
			\toprule
			\multirow{2}{*}{Method} & \multicolumn{2}{l}{Normal} & \multicolumn{2}{l}{TP} & \multicolumn{2}{l}{SD} & \multicolumn{2}{l}{MH} & \multicolumn{2}{l}{ECFB} & \multicolumn{2}{l}{Average}\\
			\cmidrule(r){2-3} \cmidrule(r){4-5} \cmidrule(r){6-7}\cmidrule(r){8-9}\cmidrule(r){10-11}\cmidrule(r){12-13}
			&  Precision      &  Jaccard  &  Precision      &  Jaccard   &  Precision      &  Jaccard    &  Precision      &  Jaccard  &  Precision      &  Jaccard &  Precision      &  Jaccard\\
			\midrule
			Gong et al. ~\cite{b38} &99.3 &0.95 &98.9 &0.96 &99.1 &0.95 &99.1 &0.94 &98.9 &0.95 &99.1 &0.95\\
			OURS    &99.6 &0.97 &99.4 &0.98 &99.5 &0.97 &99.3 &0.94 &99.3 &0.97 &99.4 &0.97\\
			\bottomrule
		\end{tabular}
	}
	\label{tab3}
\end{table*}

\begin{figure}[t]
\begin{center}
\includegraphics[width=0.75\linewidth]{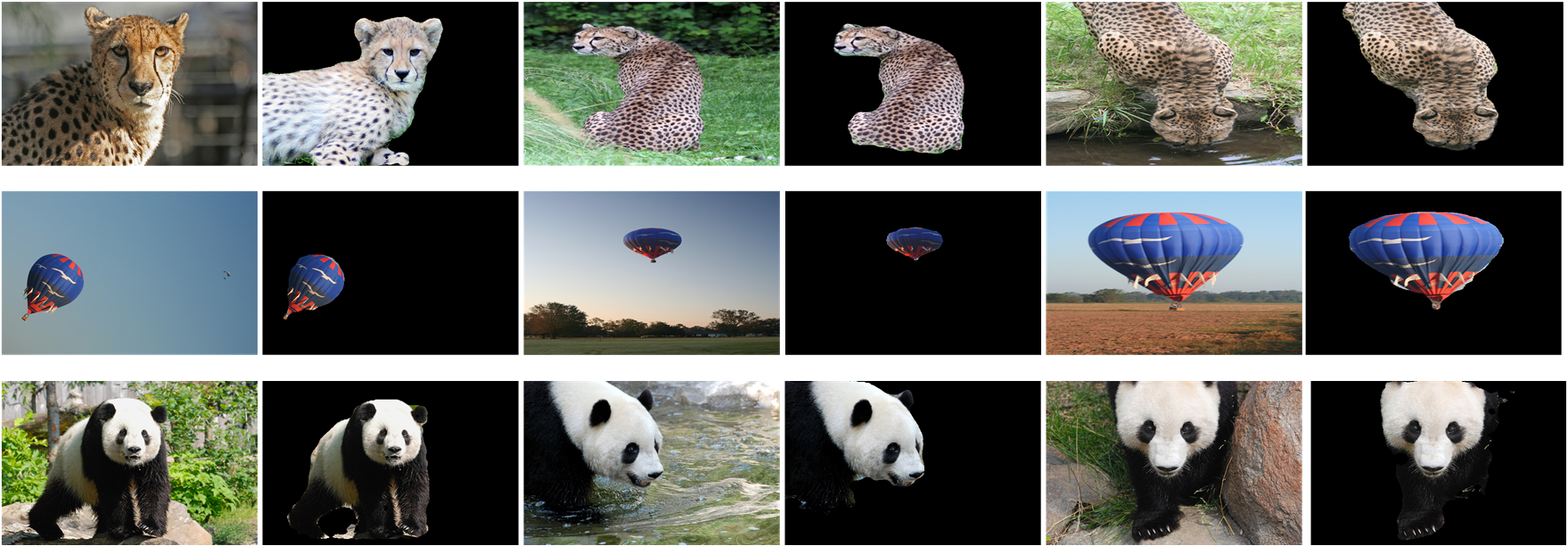}
\caption{The co-segmentation results generated by our approach on the subset of iCoseg dataset.}
\label{fig6}
\end{center}
\end{figure}

\subsection{Comparison to the State-of-the-Arts}
The performance of our image co-segmentation network is tested and compared with other seven state-of-the-art techniques including Faktor and Irani ~\cite{b35}, Jerripothula et al.~\cite{b8}, Han et al.~\cite{b5}, Yuan et al. ~\cite{b37}, Li et al.~\cite{b9}, Chen  et al.~\cite{chen2019show} and Gong  et al.~\cite{b38}. These compared methods include the conventional methods and the most recent deep learning based methods. 

The resultant performances on the Internet dataset from our method (denoted by OURS) as well as the compared methods are listed in Table~\ref{tab1}. We do ten times of tests by using our method and the value in Table~\ref{tab1} is the mean of ten tests. From the results shown in Table~\ref{tab1}, we can conclude that our method outperforms the currently best methods, not only methods based on deep neutral network but also traditional ones. It achieves the best performance on airplane and car categories on both Jaccard and Precision index, and the second best performance on horse category in terms of Precision index and the best performance in terms of Jaccard index. Specially, compare with the previous best method (from Gong et al.~\cite{b26}), the increased rates in average precision and average Jaccard index brought by our method are 1.2$\%$ and 7.9$\%$, respectively. 

Fig.~\ref{fig5} shows some examples of segmented images from our methods. As can be seen, our method can generate promising object segments under different types of intra-class variations, such as colors, sharps, views, scales and backgrounds.

To further evaluate the proposed method, we also test our approach on the subset of iCoseg, which includes eight groups of images: bear2, brown bear, cheetah, elephant, helicopter, hotballoon, panda1 and panda2. Table~\ref{tab2} shows the comparison result for each group in term of Jaccard. The results show that our method get the best performance for 5 out of 8 object groups, and it is the best one on average.

Fig.~\ref{fig6} shows some examples of co-segmentation results on iCoseg dataset by using our approach. We can see that our method accurately segments the interested objects.

\subsection{Performance on commodity images}
\begin{figure}[t]
\begin{center}
\includegraphics[width=0.75\linewidth]{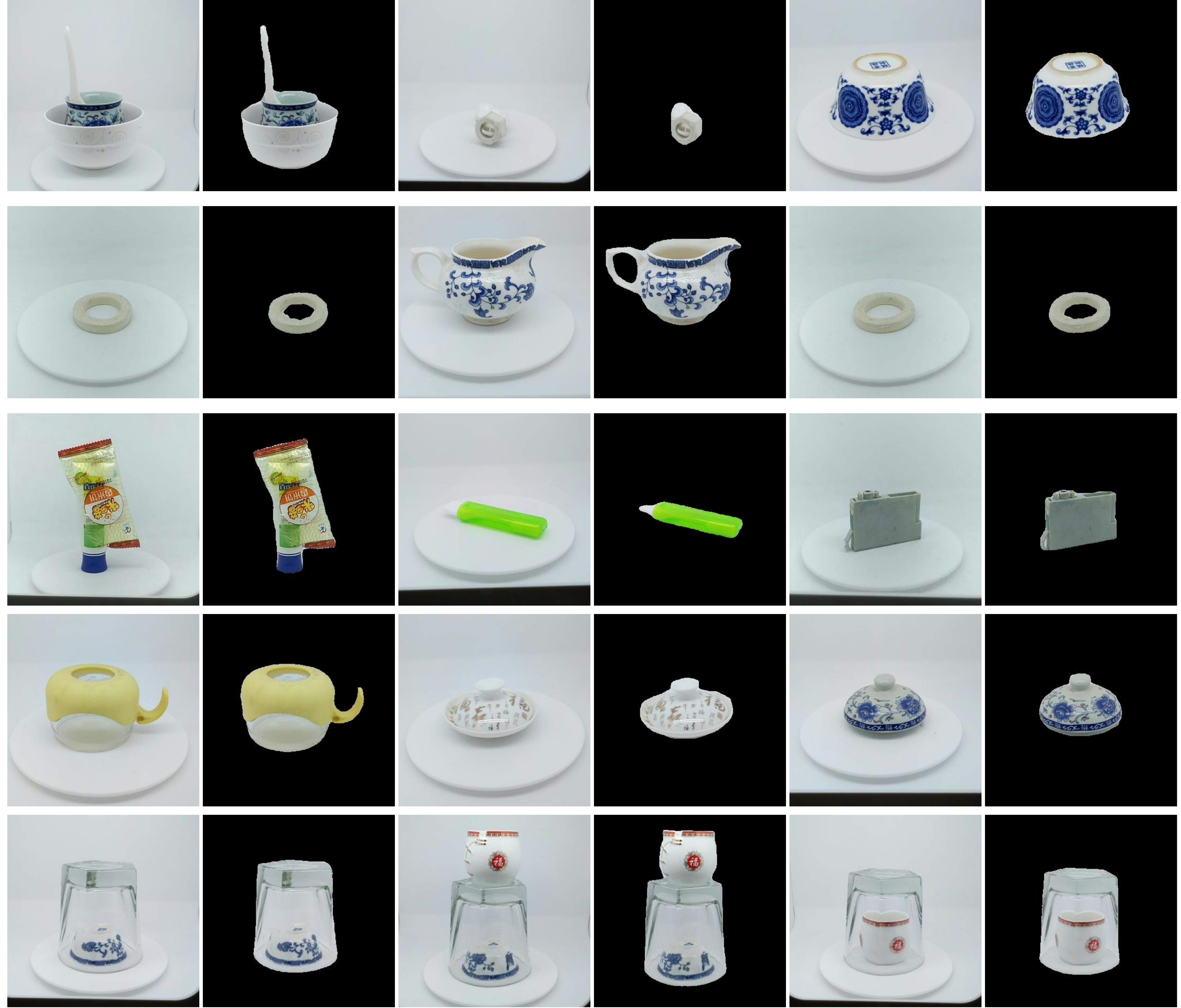}
\caption{The co-segmentation results generated by our approach on the test data of SBCoseg dataset, where the 1st, 3rd, 5th shows origins and the 2nd, 4th, 6th column shows the segmentation results; the images from top to bottom correspond to ECFB, MH, Normal, SD, and TP, respectively.}
\label{fig7}
\end{center}
\end{figure}

We use finely annotated training set and validation set to train and valid the network respectively, then test the final model on the test set. As this is the first work time to use SBCoseg dataset as the training set to train the network, so there is only comparison between our results and Gong et al. ~\cite{b38}. We show the overall performance of our approach and the performance for Normal, TP, SD, MH, and ECFP, respectively, in Table~\ref{tab3}. We then randomly choose some examples from the test set and visualize the segmented results in Fig.~\ref{fig7}. It can be seen that our network segments these five cases very well.

\subsection{Ablation study}
\begin{table}[t]
\caption{The comparisons of ablated methods on Internet, where the bold indicates the best results among all methods.}
\begin{center}
\smallskip\begin{tabular}{lll}
			\toprule
 Method& Precision& Jaccard\\
\midrule
Baseline& 94.5	& 0.76\\
Baseline+& 94.7& 0.78\\
Full& 95.7& 0.82\\
\bottomrule
\end{tabular}
\end{center}

\label{tab4}
\end{table}

Two contributions of ours in this paper are dense connection and edge enhanced 3D IOU loss of SNDM. We justify the effectiveness of them through ablation experiments in this subsection. To this purpose, we make the following changes to our image co-segmentation approach and compare the performance of them:

1) Baseline: the traditional Siamese U-net for image co-segmentation, which is same as that reported in Gong et al.~\cite{b38}. So, there is no any use of our contributions in the baseline network. It doesn’t include dense connections and is trained based on traditional IOU loss. 

2) Baseline+: We add the dense connections into the traditional Siamese U-net, but this network still outputs binary segmentation masks and is trained  by using traditional Dice loss. 

3) Full: This is the full approach of ours, including dense connection and edge enhanced 3D IOU loss of SNDM. 
We repeat the experiments ten times on the Internet with each of the above methods and record the mean performance. The training sets are still VOC 2012 and MSRC. The performance of these modified versions of our network is shown in Table~\ref{tab4}, which demonstrates that both the dense connection and the 3D IOU loss of SNDM are useful. 

\section{Conclusions}

This paper has proposed a new approach to image co-segmentation through introducing the dense connections into the decoder path of Siamese U-net and presenting a new 3D IOU loss measured over distance maps. It behaves well in the experiments. To our knowledge, the best performance on Internet and iCoseg subset is obtained by using our approach. Furthermore, the ablation study demonstrates that both dense connection and 3D IOU loss are valuable.

\section*{Acknowledgements}
This research did not receive any specific grant from funding agencies in the public, commercial, or not-for-profit sectors.


\bibliographystyle{elsarticle-num} 
\bibliography{egbib}



\end{document}